\documentclass[sigconf]{aamas} 

\usepackage{balance} 

\usepackage{algorithm}
\usepackage{algpseudocode}   

\usepackage{amsmath, amssymb}
\usepackage{mathtools}

\setcopyright{ifaamas}

\acmConference[arXiv]{arXiv preprint, under review}{United States}
{Hemanath Arumugam, Falong Fan, Bo liu}{University of Arizona}

\copyrightyear{2026}
\acmYear{2026}
\acmDOI{}
\acmPrice{}
\acmISBN{}
\usepackage{mathtools}
\mathtoolsset{showonlyrefs}
\usepackage{wrapfig}
\usepackage{amsmath}

\acmSubmissionID{595}

\title[AAMAS-2026 Formatting Instructions]{Enhancing Hierarchical Reinforcement Learning through Change Point Detection in Time Series}

\author{Hemanath Arumugam}
\affiliation{
  \institution{University of Arizona}
  \city{Tucson}
  \country{United States}}
\email{hemanath@arizona.edu}

\author{Falong Fan}
\affiliation{
  \institution{University of Arizona}
  \city{Tucson}
  \country{United States}}
\email{falongfan@arizona.edu}

\author{Bo Liu}
\affiliation{
  \institution{University of Arizona}
  \city{Tucson}
  \country{United States}}
\email{boliu@arizona.edu}

\begin{abstract}
Hierarchical Reinforcement Learning (HRL) enhances the scalability of decision-making in long-horizon tasks by introducing temporal abstraction through options—policies that span multiple timesteps. Despite its theoretical appeal, the practical implementation of HRL suffers from the challenge of autonomously discovering semantically meaningful subgoals and learning optimal option termination boundaries. This paper introduces a novel architecture that integrates a self-supervised, Transformer-based Change Point Detection (CPD) module into the Option-Critic framework, enabling adaptive segmentation of state trajectories and the discovery of options. The CPD module is trained using heuristic pseudo-labels derived from intrinsic signals—such as prediction error spikes and abrupt reward changes—to infer latent shifts in environment dynamics without external supervision. These inferred change-points are leveraged in three critical ways: (i) to serve as supervisory signals for stabilizing termination function gradients ($\beta\omega$), (ii) to pretrain intra-option policies  ($\pi\omega$) via segment-wise behavioral cloning, and (iii) to enforce functional specialization through inter-option divergence penalties over CPD-defined state partitions. The overall optimization objective enhances the standard actor-critic loss by incorporating structure-aware auxiliary losses, including binary cross-entropy for termination alignment and KL-divergence regularization for policy diversity. In our framework, option discovery arises naturally as CPD-defined trajectory segments are mapped to distinct intra-option policies, enabling the agent to autonomously partition its behavior into reusable, semantically meaningful skills. We empirically evaluate our CPD-Option-Critic framework across two benchmark domains: Four-Rooms, a discrete navigation environment requiring abstract spatial reasoning, and Pinball, a continuous control domain with sparse and delayed rewards. Results demonstrate that CPD-guided agents exhibit accelerated convergence, higher cumulative returns, and significantly improved option specialization compared to baseline Option-Critic and deliberation-cost variants. These findings confirm that integrating structural priors via change-point segmentation leads to more interpretable, sample-efficient, and robust hierarchical policies in complex environments. 

\end{abstract}


\newcommand{\BibTeX}{\rm B\kern-.05em{\sc i\kern-.025em b}\kern-.08em\TeX}

\begin{document}


\pagestyle{fancy}
\fancyhead{}


\settopmatter{printacmref=false}  
\setcopyright{none}               

\maketitle 


\section{Introduction}

Hierarchical Reinforcement Learning (HRL) offers a principled approach to scaling reinforcement learning by decomposing long-horizon tasks into subtasks, or "options"—temporally extended actions defined by initiation conditions, policies, and termination criteria. 
Among HRL methods, the Option-Critic architecture is a compelling framework that learns option policies and terminations in an end-to-end manner. Despite its flexibility, Option-Critic often struggles with two major issues: (1) termination functions suffer from sparse or vanishing gradients, leading to trivial or non-informative option boundaries, and (2) multiple options tend to collapse into redundant behaviors due to weak inductive biases on specialization. To address these, we propose augmenting the Option-Critic architecture with external segmentation signals derived from change point detection (CPD) in state trajectories. Change points naturally reflect shifts in dynamic task phases, making them ideal anchors for option terminations. We integrate Transformer-based CPD to detect these transitions and use the results to guide option learning through pretraining, regularization, and supervised alignment. 


\subsection{Hierarchical Reinforcement Learning}

By introducing \emph{temporal abstraction}, Hierarchical Reinforcement Learning (HRL) builds upon standard Reinforcement Learning (RL) and enables agents to solve complex tasks through multi-level decision-making.  HRL uses \emph{options}—temporally extended actions specified by an intra-option policy $\pi_{\omega}(a \mid s)$ and a termination function $\beta_{\omega}(s)$—instead of a single flat policy.  Effective exploration and subpolicy reuse are made possible by a high-level policy $\pi_{\Omega}(\omega \mid s)$ that decides which option to execute.

 For learning these elements straight from experience, the \emph{Option-Critic Architecture}~\cite{bacon2017optioncritic} offers a cohesive, differentiable framework.  It simultaneously optimizes option selection, intra-option control, and termination by using the \emph{Termination Gradient} and \emph{Intra-Option Policy Gradient} theorems.  For scalable and end-to-end hierarchical behavior learning, Option-Critic remains a fundamental method by fusing temporal abstraction with gradient-based optimization.


\subsection{Motivation for Integration of CPD}

The underlying transition dynamics and reward structures may change unexpectedly in dynamic or nonstationary situations, making previously learned behaviors suboptimal. The original Option-Critic architecture assumes a stationary setting and lacks mechanisms to detect such regime changes. By integrating CPD, particularly self-supervised, Transformer-based models into the Option-Critic framework, agents can monitor state trajectories or reward signals for statistically significant distributional shifts. By employing these observed change points as temporal constraints for subgoal re-evaluation, the agent can adaptively restructure or terminate choices whose behaviors are no longer consistent with the present environment. By concentrating learning updates on new or modified state regimes, this method enhances sample efficiency, improves termination sensitivity, and facilitates dynamic option discovery. Moreover, CPD-triggered option switching provides a structural prior that enhances behavioral diversity and robustness in sparse-reward or abrupt-transition domains.

\subsection{Key Contributions}

The primary contribution of this work is to integrate a Transformer-based CPD module into the Option-Critic architecture, enabling dynamic subgoal discovery and adaptive option switching. Unlike traditional HRL approaches that rely on heuristic-based or fixed segmentation strategies, our method leverages a self-supervised CPD model to identify semantically meaningful transitions in state trajectories. The detected change points are then used to enhance hierarchical learning through the following mechanisms:

\begin{itemize}
    \item \textbf{Supervised option termination:} CPD-derived change points provide direct supervision signals for learning termination functions $\beta_{\omega}(s)$, thereby stabilizing gradients and improving option switching behavior.
    
    \item \textbf{Guided intra-option policy initialization:} CPD-identified segment-level trajectories are used for behavioral cloning to initialize intra-option policies $\pi_{\omega}(a \mid s)$, accelerating early-stage learning and promoting meaningful skill specialization.
    
    \item \textbf{Regularized option diversity:} A divergence regularizer across CPD segments promotes interoption diversity and prevents collapse.
\end{itemize}

By injecting structural signals derived from CPD into the option learning loop, our architecture addresses issues such as weak termination gradients and option collapse in Option-Critic. This novel integration improves learning stability, accelerates convergence, and enhances robustness in both discrete and continuous domains.


\section{Related Work}

\subsection{Option-Critic Framework}

Bacon et al. \cite{bacon2017optioncritic} introduced the Option-Critic architecture, a key technique for end-to-end learning of choices in reinforcement learning. In this framework, an agent simultaneously learns intra-option policies (the behavior within each option) and termination functions (when options complete) using gradient-based optimization \cite{sutton1999between}. Formally, the Option-Critic derives policy gradient updates for both the option policy and its termination. In this framework, each option $\omega$ is defined by an intra-option policy $\pi_{\omega}(a \mid s)$ and a termination function $\beta_{\omega}(s)$. The parameters of the intra-option policies are updated via the \textit{Intra-Option Policy Gradient Theorem}~\cite{bacon2017optioncritic}:
\[
\nabla_{\theta} J = \mathbb{E}\left[ \nabla_{\theta} \log \pi_{\omega}(a \mid s) \; Q_U(s, \omega, a) \right]
\]
where $Q_U(s, \omega, a)$ denotes the action-value given option $\omega$. This upgrade promotes the choice to prioritize behaviors that result in greater long-term benefit (the 'better primitives' within the option). The termination function is updated according to the \textit{Termination Gradient Theorem}~\cite{bacon2017optioncritic}:
\[
\nabla_{\nu} J = \mathbb{E}\left[ -\frac{\partial \beta_{\omega}(s)}{\partial \nu} \; A^{\pi_{\Omega}}(s, \omega) \right]
\]
where $A^{\pi_{\Omega}}(s, \omega) = Q^{\pi_{\Omega}}(s, \omega) - V^{\pi_{\Omega}}(s)$ is the advantage of continuing the current option over switching ~\cite{sutton1999between}. This update, intuitively, lengthens advantageous choices by reducing the termination probability when an option has a positive benefit. In conclusion, the Option-Critic framework provides a stochastic actor-critic approach for HRL that learns choices solely from reward signals, thereby eliminating the need for resets or pre-established subgoals. Numerous subsequent extensions in option discovery were made possible by it \cite{harb2018deliberation}.

\subsection{Subgoal Discovery in Hierarchical RL}

Identifying effective subgoals or bottleneck states that choices should automatically reach is a significant challenge in HRL. Recent research focuses on identifying subgoals from data, whereas earlier methods frequently assumed they were provided. For instance, Kulkarni et al. ~\cite{kulkarni2016hierarchical} introduced a two-level hierarchy, Hierarchical-DQN (h-DQN), in which a low-level DQN chases these subgoal states with intrinsic incentives, while a high-level policy chooses objectives (defined as certain landmark states). h-DQN significantly enhanced exploration in tasks such as Montezuma's Revenge, where obtaining specific subgoal objects (keys, doors, etc.) is crucial, by utilizing expert-provided goal pixels for termination conditions. Beyond manually defined goals, other methods learn subgoals by analyzing agent experience. A common strategy is to identify bottleneck states – states that lie on many successful trajectories but are seldom visited in unsuccessful ones. Such states (e.g., doorways connecting rooms in a maze) serve as natural subgoals. Algorithms detect these by comparing state visitation frequencies between successful and failed trials. Reaching a bottleneck state can be set as an option’s objective, since these states grant access to new regions of the state space. Other approaches rely on graph partitioning of the state space: for instance, the Q-Cut algorithm models the environment as a graph. It identifies cut points (edges) whose removal partitions the graph, suggesting subgoal states on the boundaries. Overall, subgoal discovery techniques enrich HRL by proposing meaningful intermediate targets that accelerate exploration and learning in long-horizon tasks~\cite{machado2017laplacian,daniel2016probabilistic,gregor2016vic,riemer2018abstract,fox2017multilevel,sharma2017optiondiscovery}.

\subsection{Deliberation Cost}

The Option-Critic architecture often suffers from premature option termination, where options collapse into single-step actions. Harb et al.~\cite{harb2018deliberation} addressed this issue by introducing a \emph{deliberation cost}, a small penalty $-\eta$ applied whenever an option terminates. This adjusts the termination gradient by modifying the advantage term~\cite{harb2018deliberation}:
\[
A'(s, \omega) = A(s, \omega) - \eta,
\]
so that an option switches only when another yields at least $\eta$ higher return. This mechanism discourages trivial switching and promotes temporally extended behaviors. The deliberation cost embodies a bounded rationality principle—agents change strategies only when the change is clearly beneficial—and has been adopted in later variants, such as A2OC, to improve option duration, specialization, and stability across both discrete and continuous domains.

\subsection{Statistical and Model-Based CPD}

The goal of change point detection (CPD) is to locate time-series data points where underlying statistical characteristics change. The Bayesian Online Change Point Detection (BOCPD) framework by Adams and MacKay~\cite{adams2007bocpd} is a fundamental model-based method that updates a posterior iteratively with each observation and maintains it across the run length $r_t$, or the period since the last change. The hazard function $H(r)$ governs the likelihood of starting a new segment, and the model assumes that data are created from piecewise i.i.d.\ segments. BOCPD generates a probability of change at each timestep by computing the joint posterior $P(r_t, x_{1:t})$ online via a message-passing update.

A key strength of Bayesian CPD lies in its modularity: different likelihood models (e.g., Gaussian, Poisson) can be embedded to detect shifts in underlying generative parameters. This enables real-time, probabilistic segmentation of non-stationary sequences and robust detection of regime changes across domains such as finance and biometrics~\cite{truong2020review,aminikhanghahi2017survey,chen2012changepoint}.

\subsection{Learning-Based CPD}

Recent studies investigate learning-based change point detection (CPD) approaches that utilize deep and self-supervised learning, in addition to conventional model-driven approaches, to detect changes without making explicit parametric assumptions. These methods learn representations that can capture complex and high-dimensional transitions. For example, Deldari et al.~\cite{deldari2021time} introduced TS-CP$^2$, a self-supervised contrastive framework that distinguishes adjacent segments from those separated by change points. Other architectures, such as autoencoders and recurrent models, detect shifts through spikes in reconstruction or prediction error. In reinforcement learning, similar intrinsic signals, such as reward prediction error or surprise, can serve as pseudo-labels for training neural CPD detectors. This concept is further refined by hybrid models such as CLaSP and PaTSS, which categorize whether segment borders exist. Learning-based CPD often performs better than traditional methods on modern time-series benchmarks because it relaxes distributional assumptions and uses high-level representations to capture subtle, context-dependent changes ~\cite{vaswani2017attention,zhou2021informer,nie2022timeseries,wu2021autoformer}. To ensure successful generalization, these techniques require domain-informed pseudo-labels or meticulous task design.

\subsection{CPD in Reinforcement Learning}

In reinforcement learning (RL), CPD has been increasingly used to manage non-stationary settings and understand agent behavior. When transition models or reward functions change, CPD enables the system to recognize these shifts and adjust its policies accordingly. Early work by Hadoux et al.~\cite{hadoux2014sequential} applied a sequential, CUSUM-based CPD method to identify environment shifts, while Alegre et al.~\cite{alegre2021minimum} later proposed a real-time adaptation algorithm that minimizes detection delay and false alarms. In addition to supporting rapid adaptation, CPD has been applied to segment trajectories into behavioral phases—such as exploration and exploitation—offering interpretable insights into learning progression and informing hierarchical policy design~\cite{xie2020cpdrl,le2018multiagent}.

\subsection{Structure-Aware Option Learning}

A central goal in HRL is \emph{structure-aware} option discovery—learning options that align with meaningful behavioral segments in the state space. Ranchod et al.~\cite{ranchod2015bayesian} used a nonparametric Bayesian model to segment demonstrations into skill phases via inferred reward functions. In contrast, Cockcroft et al.~\cite{cockcroft2020nbprs} mapped these segments into options whose boundaries define initiation and termination sets. Such segmentation ensures each option captures a coherent subpolicy rather than arbitrary behavior fragments.

In reinforcement learning without demonstrations, CPD offers an unsupervised signal for discovering option boundaries. Sudden shifts in dynamics or prediction errors can guide the termination function $\beta(s)$ and provide segmented trajectories for intra-option policy learning. This aligns option definitions with empirically observed behavioral regimes.

Similar in spirit to the termination critic~\cite{harutyunyan2019termination}, which encourages compact termination regions, CPD-based supervision grounds option boundaries in data-driven temporal structure rather than reward heuristics~\cite{bacon2017optioncritic,harb2018deliberation}. By integrating CPD within the option-learning process, agents can form interpretable, reusable options that mirror latent task phases, enhancing both learning efficiency and policy clarity.

\makeatletter
\renewcommand{\theequation}{\arabic{equation}}
\makeatother
\setcounter{equation}{0}

\section{Our contributions}

Our proposed framework enhances the standard Option-Critic architecture by incorporating a Transformer-based Change Point Detection (CPD) model that provides structural guidance during learning. This integration introduces three major innovations: (i) CPD-supervised termination learning, (ii) CPD-guided option pretraining, and (iii) inter-option policy diversity enforcement. Each of these components is formalized below.

\subsection{CPD-Supervised Termination Learning}

With gradients determined by the termination advantage~\eqref{eq:aterm}, the termination function is learnt by optimizing the expected return in the original Option-Critic framework. Following the Option-Critic framework~\cite{bacon2017optioncritic}, the termination advantage is defined as

\begin{equation}\label{eq:aterm}
A_{\mathrm{term}}(s) = Q(s, \omega) - V(s)
\end{equation}

However, particularly in the early phases of learning, these gradients, such as ~\eqref{eq:aterm}, are frequently noisy or sparse. We utilize the CPD model as a source of pseudo-labels and propose an auxiliary binary classification loss to stabilize this process. The change point detector identifies time steps where the state distribution undergoes a significant shift. These points are treated as ground truth labels for learning termination boundaries.

We define the termination loss as:
\begin{equation}\label{eq:tloss}
\mathcal{L}_{\mathrm{term}} = \mathrm{BCE}\big(\beta_{\omega}(s), y_{\mathrm{CPD}}(s)\big)
\end{equation}

where
\begin{equation}\label{eq:bloss}
\mathrm{BCE}(p, y) = -y\log(p) - (1 - y)\log(1 - p)
\end{equation}

and $y_{\mathrm{CPD}}(s) \in \{0,1\}$ is the CPD output indicating a change point at state $s$, $\beta_{\omega}(s) \in [0,1]$ is the learned termination probability for option~$\omega$ from equation~\eqref{eq:bloss}. By serving as a regularizer, this supplementary loss ~\eqref{eq:tloss} grounds termination decisions in the environment's CPD-inferred structure.




\subsection{CPD-Guided Option Pretraining via Behavioral Cloning}

To accelerate option discovery, we segment the full trajectories of $\{\tau_i\}$ using CPD into temporally coherent segments, each reflecting a distinct behavioral phase. Each segment is assigned a unique option $\omega_i$, and intra-option policies $\pi_{\omega_i}(a \mid s)$ are pretrained using behavioral cloning:
\begin{equation}\label{eq:bcl}
\mathcal{L}_{\mathrm{BC}} = - \sum_{(s,a) \in \tau_i} \log \pi_{\omega_i}(a \mid s)
\end{equation}
This imitation loss encourages each option to reproduce the behavior observed within its assigned segment. This initialization biases intra-option policies toward structure-informed specialization, aligning with latent task phases.

\subsection{Option Diversity via Segment-Wise Separation}

To avoid option collapse, we introduce a diversity regularization term. Using CPD-defined segments as state partitions, we enforce different options to exhibit divergent action distributions over distinct subspaces. We define the diversity loss as:
\begin{equation}\label{eq:ldiv}
\mathcal{L}_{\mathrm{div}} = \sum_{\omega_i \neq \omega_j} D_{\mathrm{KL}}\big(\pi_{\omega_i}(\cdot \mid s) \,\|\, \pi_{\omega_j}(\cdot \mid s)\big) \quad \text{for} \quad s \in S_{\mathrm{CPD}}
\end{equation}

where: $D_{\mathrm{KL}}$ is the KL-divergence, and $S_{\mathrm{CPD}}$ is the union of all disjoint state subspaces defined by CPD segments from equation~\eqref{eq:ldiv}. This regularizer promotes functional specialization of options by penalizing redundancy.



\subsection{Combined Training Objective}

The total loss function guiding the optimization of the CPD-Option-Critic agent is a weighted sum of the RL loss and auxiliary components:

\begin{equation}\label{eq:lt}
\mathcal{L}_{\mathrm{total}} = \mathcal{L}_{\mathrm{RL}} + \alpha \cdot \mathcal{L}_{\mathrm{term}} + \beta \cdot \mathcal{L}_{\mathrm{div}} + \gamma \cdot \mathcal{L}_{\mathrm{BC}}
\end{equation}


where $\mathcal{L}_{\mathrm{RL}}$ is the standard actor-critic loss that follows the objective defined in the original Option-Critic architecture~\cite{bacon2017optioncritic}, $\mathcal{L}_{\mathrm{term}}$ serves as the CPD-supervised termination loss~\eqref{eq:bloss}; $\mathcal{L}_{\mathrm{div}}$ is the KL-divergence-based diversity loss, and $\alpha, \beta, \gamma$ are weighting hyperparameters. 

By embedding structural knowledge derived from CPD directly into the learning dynamics, our framework encourages temporally coherent option terminations, meaningful intra-option behavior, and robust exploration. This principled integration of segmentation-aware supervision with the Option-Critic design, including the total loss functions~\eqref{eq:lt}, significantly improves learning speed, option interpretability, and adaptability to non-stationary dynamics.




\section{Architecture and Algorithms}

\begin{figure}
    \centering
    \includegraphics[width=\linewidth]{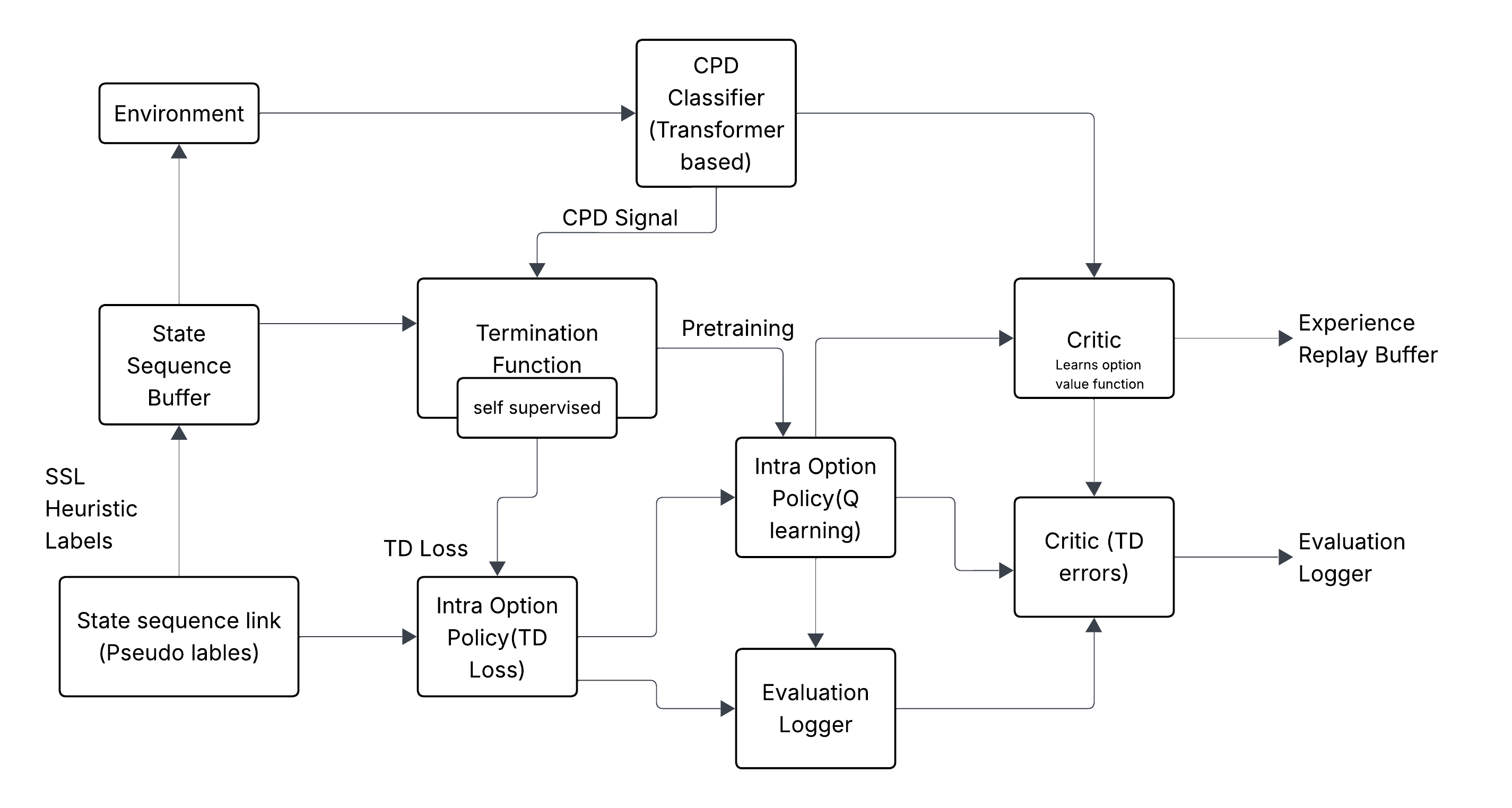}
    \caption{\footnotesize Workflow of CPD Integrated Option Critic framework}
    \label{fig:architecture}
\end{figure}

The proposed architecture extends the traditional Option-Critic framework by embedding a self-supervised CPD module, enabling autonomous and temporally coherent option discovery in reinforcement learning. The agent interacts with the environment to collect transitions $(s_t, a_t, r_t, s_{t+1})$, which are stored in a state sequence buffer. These trajectories are processed by a Transformer-based CPD classifier trained using pseudo-labels generated from self-supervised heuristics such as TD-error spikes or sudden reward shifts, which approximate latent structural transitions. The CPD classifier minimizes a binary cross-entropy loss:
\begin{equation}\label{eq:bcpd}
\mathcal{L}_{\text{CPD}} = \mathbb{E}_{(s_{1:T})} \left[ -\sum_{t=1}^{T} y_t \log \hat{y}_t + (1 - y_t) \log (1 - \hat{y}_t) \right]
\end{equation}
where $y_t \in \{0, 1\}$ is the pseudo-label indicating change points and $\hat{y}_t \in [0,1]$ is the predicted probability from the classifier mentioned in ~\eqref{eq:bcpd}

The resulting CPD signals serve two critical roles: (i) they supervise the termination function $\beta_\omega(s) \in [0, 1]$, enhancing the standard advantage-based termination gradient with a regularization term:
\begin{equation}\label{eq:atg}
\mathcal{L}_{\text{term}} = \mathcal{L}_{A2O} + \lambda_{\text{CPD}} \cdot \mathcal{L}_{\text{CPD}}(\beta_\omega(s), y_{\text{CPD}}),
\end{equation}
and (ii) they segment trajectories into behaviorally distinct intervals used to initialize intra-option policies $\pi_\omega(a|s)$ via behavioral cloning:
\begin{equation}\label{eq:lbc}
\mathcal{L}_{\text{BC}} = -\mathbb{E}_{(s, a)} [\log \pi_\omega(a|s)],
\end{equation}
which accelerates early-stage specialization.

These intra-option policies are further refined using Q-learning and TD loss, guided by critic updates:

\begin{align}
Q_\Omega(s, \omega) &= \mathbb{E}_{a \sim \pi_\omega} \left[ r(s,a) + \gamma V_\Omega(s') \right], \\
V_\Omega(s) &= \sum_\omega \pi_\Omega(\omega|s) Q_\Omega(s, \omega),
\end{align}
\begin{equation}
\mathcal{L}_{\text{TD}} = \left( r_t + \gamma Q_\Omega(s_{t+1}, \omega_t) - Q_\Omega(s_t, \omega_t) \right)^2.
\end{equation}

The critic estimates both the option-value $Q_\Omega(s, \omega)$ and the state-value $V(s)$, which support computation of the option advantage $A(s, \omega) = Q_\Omega(s, \omega) - V(s)$ ~\cite{bacon2017optioncritic} used for updating both intra-option policies and termination probabilities. To mitigate option redundancy and enforce behavioral diversity, a KL-divergence-based regularization term is included:
\begin{equation}\label{eq:ldivt}
\mathcal{L}_{\text{div}} = \sum_{\omega \ne \omega'} D_{\text{KL}} \left( \pi_\omega(\cdot|s) \parallel \pi_{\omega'}(\cdot|s) \right),
\end{equation}
where $D_{\text{KL}}$ denotes the Kullback-Leibler divergence across intra-option policies conditioned on state $s$.

The total loss optimized during training thus combines all relevant signals:
\begin{equation}\label{eq:ltlop}
\mathcal{L}_{\text{total}} = \mathcal{L}_{\text{TD}} + \lambda_{\text{CPD}} \mathcal{L}_{\text{CPD}} + \lambda_{\text{BC}} \mathcal{L}_{\text{BC}} + \lambda_{\text{div}} \mathcal{L}_{\text{div}},
\end{equation}
coordinating value estimation, segmentation, option specialization, and termination. Experience tuples are stored in a replay buffer for off-policy updates, while an evaluation module logs episodic return, CPD accuracy, termination frequency, and option usage. The overall architecture, as represented in Figure \ref{fig:architecture}, retains the learning dynamics of the original Option-Critic design while significantly enhancing temporal abstraction, interpretability, and robustness through CPD-driven guidance.

\begin{algorithm}[t]
\caption{\footnotesize Transformer-Based Change Point Detection (CPD)}
\label{alg:cpd}
\begin{algorithmic}[1]
\Require Trajectory $\tau=\{(s_t,a_t,r_t)\}_{t=1}^T$, intrinsic signals $z_t$, context window $W$, threshold $\gamma$
\Ensure Boundary indicators $\hat b_t \in \{0,1\}$ and segments $\mathcal{S}$
\State Build tokens $x_t \gets \mathrm{concat}(\phi(s_t), a_t, r_t, z_t)$
\For{$t=1$ to $T$}
  \State $h_t \gets \mathrm{TransformerEncoder}(x_{t-W:t})$ \Comment{causal or bi-causal}
  \State $p_t \gets \sigma(g_\theta(h_t))$ \Comment{boundary probability}
\EndFor
\State Compute pseudo-labels $y_t \in \{0,1\}$ from peaks/changes in $z_t$ (smoothed in $\pm \Delta$ window)
\State $\tilde y_t \gets (1-\varepsilon) y_t + \frac{\varepsilon}{2}$ \Comment{label smoothing}
\State $\mathcal{L}_{\text{cpd}} \gets -\sum_t \big[(1-\tilde y_t)\log(1-p_t) + \tilde y_t \log p_t \big]$
\State (Optional) $\mathcal{L}_{\text{aux}} \gets \sum_t \lVert \hat u_{t+1}-u_{t+1} \rVert_2^2$ \Comment{e.g., forecast feature/reward deltas}
\State $\mathcal{L}_{\text{CPD-total}} \gets \lambda_{\text{sup}}\mathcal{L}_{\text{cpd}}+\lambda_{\text{aux}}\mathcal{L}_{\text{aux}}$
\State Update $\theta$ by minimizing $\mathcal{L}_{\text{CPD-total}}$
\State Infer $\{p_t\}$ on $\tau$ and set $\hat b_t \gets \mathbf{1}[p_t \ge \gamma]$
\State Form segments $\mathcal{S}$ by splitting $\tau$ at indices with $\hat b_t=1$
\State \Return $\{\hat b_t\}$, $\mathcal{S}$
\end{algorithmic}
\end{algorithm}

The CPD module learns to automatically identify moments where the agent’s interaction dynamics change significantly. 
Instead of relying on handcrafted thresholds or prior knowledge, a Transformer network observes raw trajectories composed of state, action, and reward signals. 
By training on self-supervised pseudo-labels derived from intrinsic signals, such as prediction-error spikes or abrupt reward changes, the model learns a representation in which change points naturally emerge as boundaries with high probability. 
During inference, these boundary probabilities are thresholded to segment the trajectory into coherent behavioral chunks, which later serve as the structural basis for both option termination and subgoal discovery. 
In essence, Algorithm 1 enables the agent to ``perceive'' shifts in task context directly from experience.

\begin{algorithm}[t]
\caption{\footnotesize CPD-Guided Termination Supervision for $\beta_\omega$}
\label{alg:beta}
\begin{algorithmic}[1]
\Require Option-Critic components $(\{\pi_\omega\}, \{\beta_\omega\}, Q)$, boundary probs $p_t$ (or $\hat b_t$), weights $\lambda_\beta, \lambda_{\text{CPD}}$, sharpen temperature $\tau$, boundary neighborhood $\mathcal{N}$
\Ensure Updated termination parameters $\{\beta_\omega\}$
\For{each timestep $t$ with active option $\omega_t$}
  \State $\hat\beta_t \gets \beta_{\omega_t}(s_t)$
  \State $\tilde b_t \gets \mathrm{Sharpen}(p_t;\tau)$ \Comment{soft target, higher near boundaries}
  \State $w_t \gets 1 + \alpha \cdot \mathbf{1}[t \in \mathcal{N}]$ \Comment{upweight near boundaries}
  \State Accumulate $\mathcal{L}_\beta \mathrel{+}= -w_t\big[\tilde b_t\log \hat\beta_t + (1-\tilde b_t)\log(1-\hat\beta_t)\big]$
\EndFor
\State $\mathcal{L}_{\text{total}} \gets \mathcal{L}_{\text{OC}} + \lambda_\beta \mathcal{L}_\beta + \lambda_{\text{CPD}}\mathcal{L}_{\text{CPD-total}}$
\State Update all parameters with $\nabla \mathcal{L}_{\text{total}}$ \Comment{policy, critic, termination, CPD}
\State (Optional) Warm-up: freeze $\beta_\omega$ for $K$ steps until CPD stabilizes
\State \Return $\{\beta_\omega\}$
\end{algorithmic}
\end{algorithm}

Once potential boundaries are detected, the CPD signal is used to supervise the termination function $\beta_\omega$ of each option. 
Instead of learning termination points purely from high-variance policy gradients, $\beta_\omega$ receives soft targets based on the CPD boundary probabilities. 
This coupling guides the termination network to increase its probability near genuine behavioral transitions while maintaining stability elsewhere. 
As a result, the agent learns when to switch options in a data-driven yet smooth manner, thereby avoiding premature termination or collapsing into a single option. 
Algorithm 2 effectively aligns low-level decision boundaries with the high-level temporal structure revealed by the CPD model.

\begin{algorithm}[t]
    \caption{\footnotesize Subgoal Discovery from CPD Segments with BC Pretraining}
    \label{alg:subgoals}
    \begin{algorithmic}[1]
    \Require Segments $\mathcal{S}$, encoder $\phi$, desired subgoals $K$ (or auto), bonus $\alpha$
    \Ensure Subgoals $\mathcal{G}$, initialized options $\{(\pi_{\omega_k}, \beta_{\omega_k})\}_{k=1}^K$
    \State \textbf{(A) Discover subgoals}
    \For{each segment $S_m \in \mathcal{S}$}
      \State $u_m \gets \mathrm{Pool}(\{\phi(s_t)\mid t \in S_m\})$ \Comment{mean/attentive pooling}
    \EndFor
    \State Cluster $\{u_m\}$ into $K$ groups $\{C_k\}$ with prototypes $\{c_k\}$
    \State Define subgoals $\mathcal{G} \gets \{g_k \leftarrow c_k\}_{k=1}^K$; assign option $\omega_k \leftrightarrow g_k$
    \State \textbf{(B) Behavior cloning pretraining}
    \For{$k=1$ to $K$}
      \State Build $\mathcal{D}_k \gets \{(s,a) \text{ from segments in } C_k\}$
      \State Train $\pi_{\omega_k}$ with $\mathcal{L}_{\omega_k}^{\text{BC}} = -\sum_{(s,a)\in \mathcal{D}_k} \log \pi_{\omega_k}(a\mid s)$
    \EndFor
    \State \textbf{(C) RL fine-tuning with shaping + CPD $\beta$}
    \State Initialize $\beta_{\omega_k}(s)$ higher near neighborhoods of $g_k$ and CPD boundaries
    \State Train Option-Critic with $r'_t \gets r_t + \alpha \cdot \mathbf{1}[\text{reach } g_k]$
    \State Continue Alg.~\ref{alg:beta} for CPD-guided $\beta$ updates; periodically re-estimate $\mathcal{G}$ if drift
    \State \Return $\mathcal{G}$, $\{(\pi_{\omega_k}, \beta_{\omega_k})\}$
    \end{algorithmic}
\end{algorithm}

The subgoal discovery module translates CPD-based trajectory segments into meaningful sub-tasks that each option can specialize in. 
Embeddings of the segmented state sequences are clustered to identify recurring patterns, which form subgoal prototypes representing intermediate milestones in the environment. 
Each option policy is then behavior-cloned on the corresponding cluster’s trajectories to quickly learn its characteristic behavior before being fine-tuned with reinforcement learning. 
This process ensures that options start from distinct, interpretable skills and continue to refine them through interaction. 
Together with CPD-aligned terminations, this mechanism, as mentioned in Algorithm 3, promotes structured exploration, faster convergence, and semantically coherent option behaviors.


\section{Experimental Setup}

To evaluate the effectiveness of CPD-guided option discovery, we conduct experiments in two distinct environments that pose complementary challenges in terms of exploration, state abstraction, and temporal structure: the Four-Rooms environment and the Pinball environment.

\subsection{Four-Rooms}

The Four-Rooms grid-world scenario, a common benchmark for HRL by \citet{sutton1999between}, is used to test our design. The $13 \times 13$ grid comprises four interconnected rooms (169 discrete states) separated by walls, where the agent can move in four directions: up, down, left, or right. Actions are stochastic—each intended move succeeds with a 66.7\% probability, while with 33.3\% probability a random adjacent action is executed. To address this stochasticity, all experiments are averaged across more than 100 random seeds.

The reward is sparse: the agent receives $+1.0$ only upon reaching the goal and $0.0$ otherwise. Following the setup from the original Option-Critic paper \citep{bacon2017optioncritic}, each episode terminates upon reaching the goal or after 500 steps. We introduced non-stationary goals shifting at episode 1000 testing adaptability under distributional change.

We are using 4 options and 8 options to implement two Option-Critic baselines, OC-4 and OC-8. Tabular representations with a Boltzmann policy (temperature $= 0.001$) are used by the option-value functions $Q_{\Omega}(s,\omega)$, termination functions $\beta_{\omega}(s)$, and intra-option policies $\pi_{\omega}(a|s)$.  Learning rates are $\alpha_{\mathrm{critic}} = 0.5$, $\alpha_{\theta} = 0.25$, and $\alpha_{\beta} = 0.25$. Deterministic option selection ($\epsilon_{\mathrm{option}}=0.0$) and a discount factor $\gamma = 0.99$ are also present.

\subsection{Pinball Domain}

We utilize the Pinball domain, a 2D physics-based labyrinth in which the agent (a ball) must achieve a goal while avoiding polygonal obstacles, as introduced in the original Option-Critic work \citep{bacon2017optioncritic}, to assess performance in continuous control. In the 4D state space $(x, y, v_x, v_y)$, actions are equivalent to four-directional impulses. Elastic collisions with drag ($0.995$) and sparse rewards: $+10{,}000$ for goal achievement, $-1$ per null action, $-5$ per wrong step, and episode termination after $10{,}000$ steps.

We reproduced the original Option-Critic setup with four options, neural network parameterization, and a Fourier basis of order 4 (resulting in a 256-dimensional encoding). The critic and intra-option policy learning rates are $0.001$, termination learning rate $0.0001$, and discount factor $\gamma = 0.99$. Exploration follows an $\epsilon$-greedy strategy starting at $0.1$ and decaying exponentially.

Our ChangemPoint Detection (CPD) module augments the Option-Critic framework with online segmentation and adaptive supervision. It employs a linear 64-dimensional state encoder, a 4-layer multi-head attention network (4 heads), and a sigmoid-based binary change classifier (threshold $0.6$). Using a memory buffer of 1,000 samples and trajectory windows of length 20, CPD is trained with Adam (lr $=0.001$, weight decay $=1\mathrm{e}{-4}$), gradient clipping (1.0), and Gaussian noise ($\sigma=0.01$ with 30\% probability).

\section{Results and Analysis}

\subsection{Four-Room Experiment}

\begin{figure}
    \centering
    \includegraphics[width=1\linewidth]{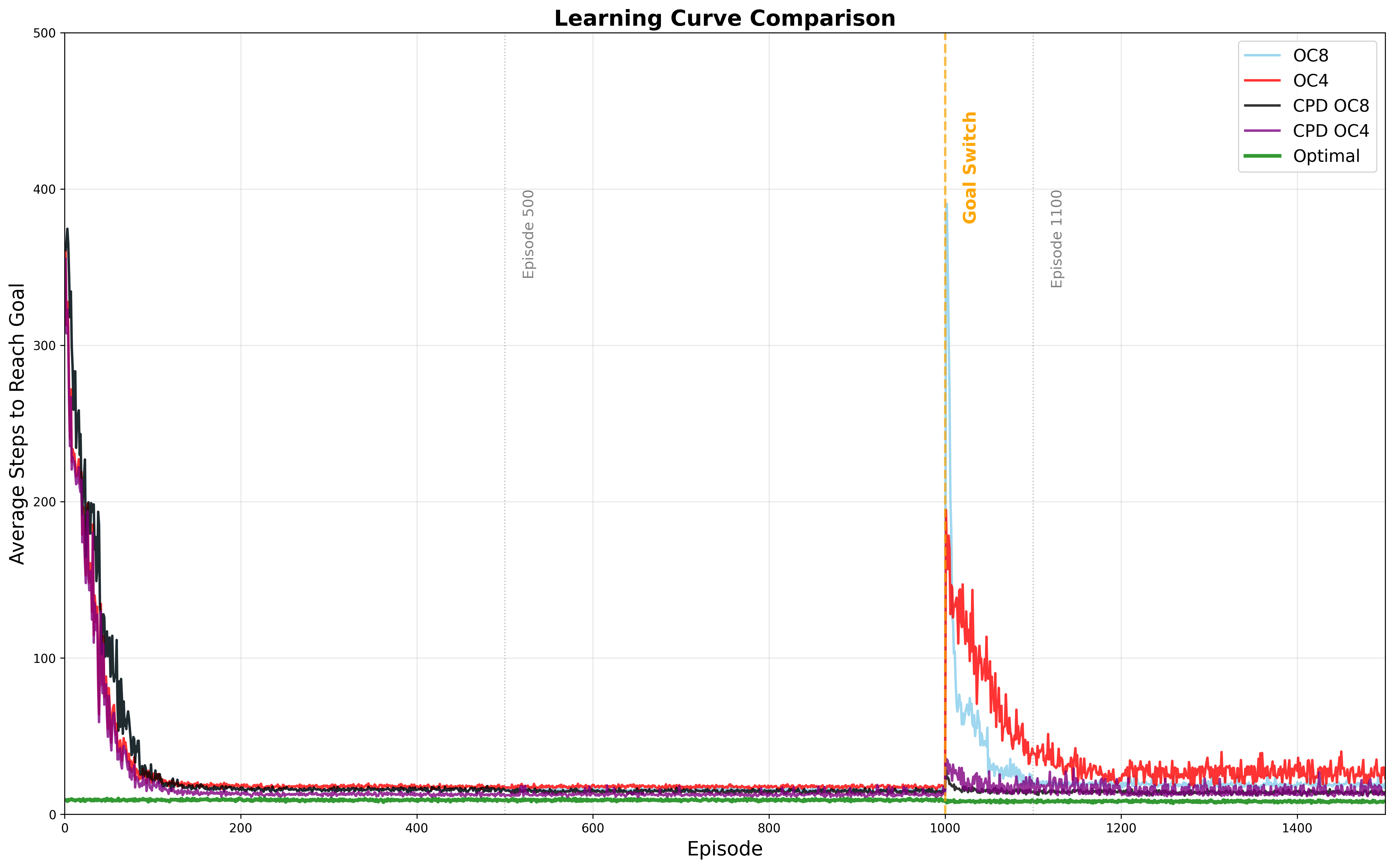}
    \caption{\footnotesize Agent's Learning Comparison Curve. }
    \label{fig:Fourroom Learning Curve}
\end{figure}

The experimental results in the Four-Rooms environment reveal a clear and consistent advantage of integrating Transformer-based CPD into the Option-Critic architecture. Across all variants—OC4, OC8, CPD-OC4, and CPD-OC8—the CPD-enhanced agents exhibit faster convergence, greater stability, and superior adaptability in response to environmental structural changes. As shown in Figure \ref{fig:Fourroom Learning Curve}, the average number of steps to reach the goal declines rapidly during the initial training phase for all agents, indicating that temporal abstraction via options accelerates early exploration. However, the distinction between baseline and CPD-augmented architectures becomes particularly evident after the goal-switch event at episode 1000, where the environment’s reward structure and state transitions are altered. While the baseline OC4 and OC8 agents experience a marked increase in average steps following the switch, both CPD variants recover much more quickly, reflecting their ability to identify environmental shifts and adapt their high-level policies through CPD-guided termination and option switching.


\begin{wrapfigure}[15]{r}{0.3\textwidth} 
    \centering
    \vspace{-10pt} 
    \includegraphics[width=0.28\textwidth]{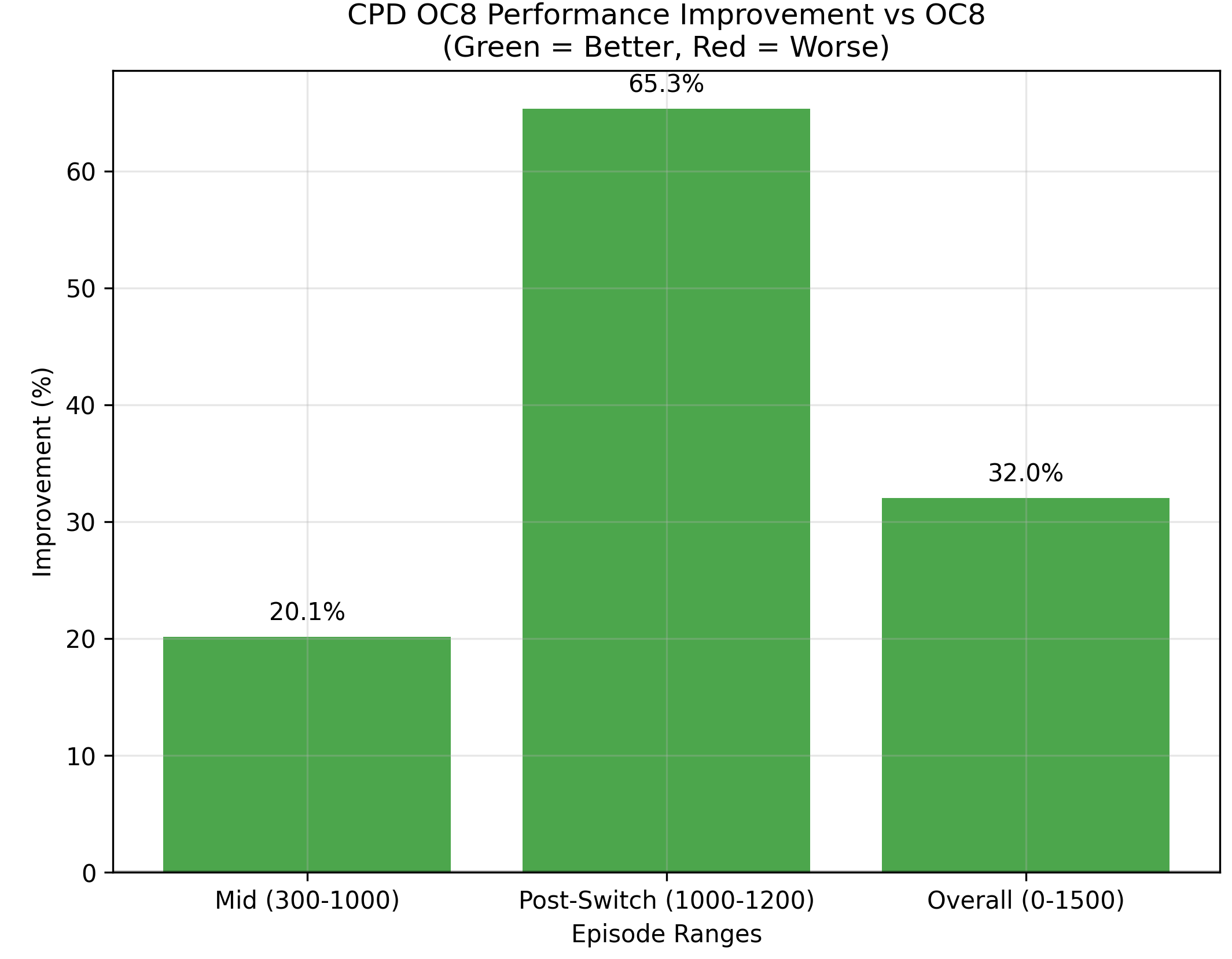}
    \caption{\footnotesize Performance comparison of CPD-enabled Option Critic with Option Critic agent using 8 options.}
    \label{fig:Performance Comparison}
    \vspace{-10pt}
\end{wrapfigure}

Quantitatively, these behavioral differences translate into substantial performance gains. As summarized in Figure \ref{fig:Performance Comparison}, the CPD-OC8 agent achieves an overall 32\% reduction in average steps per episode compared to its baseline OC8 counterpart, while the CPD-OC4 agent attains an even higher 36\% improvement relative to OC4. The post-switch period (episodes 1000–1200) shows the most significant gain, with CPD-OC8 and CPD-OC4 outperforming their baselines by approximately 65\%, highlighting their greater ability to generalize in non-stationary settings. Both CPD agents maintain approximately 20–28\% lower step counts than their baselines, even during the mid-phase of training (episodes 300–1000), when the environment is stable. This suggests that by improving the temporal segmentation of policies, CPD not only increases adaptation but also promotes long-term efficiency. That is, instead of relying solely on learning gradient cues, the transformer-based CPD module enables the agent to predict and match choice terminations with underlying structural limits, such as doorway transitions.

\begin{wrapfigure}{R}{0.3\textwidth}
    \centering
    \includegraphics[width=0.3\textwidth]{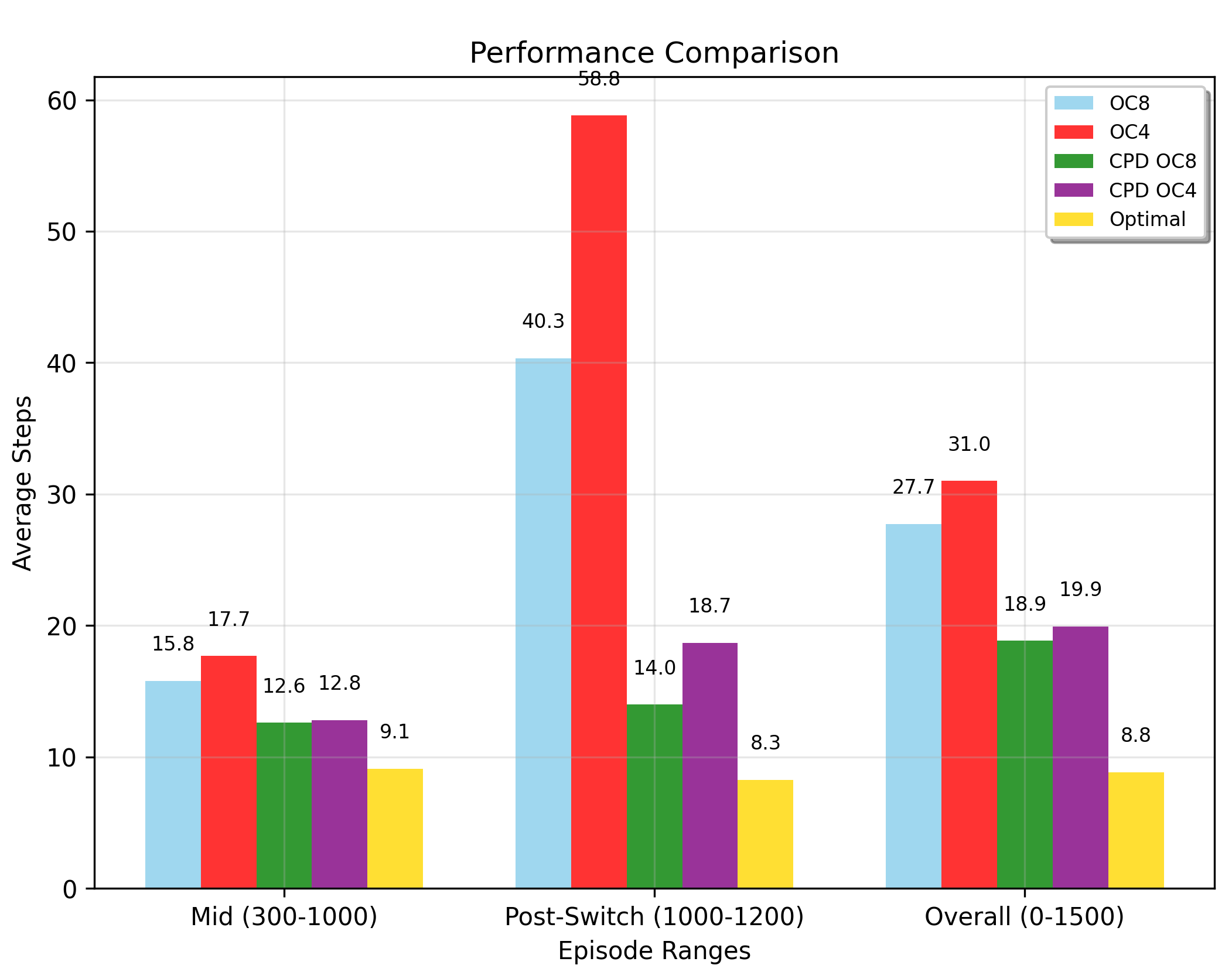}
    \caption{\footnotesize Comparison of average steps to reach the goal, including optimal steps.}
    \label{fig:Comparison}
\end{wrapfigure}

An essential element of this experiment is the inclusion of the “Optimal” trajectory baseline, as shown in Figures \ref{fig:Fourroom Learning Curve} and \ref{fig:Comparison}. This benchmark represents the theoretical minimum number of steps required for a perfectly informed agent to reach the goal—typically between 8 and 10 steps in a 13×13 Four-Rooms gridworld—assuming full observability and deterministic state transitions. It is computed analytically via A* search or dynamic programming and provides a lower bound on the achievable path length. Comparing each agent’s performance to this theoretical optimum allows us to measure path efficiency, defined as the ratio of the optimal step count to the observed average. The CPD-enhanced agents consistently maintain trajectories closer to this bound, with approximately 47\% higher efficiency than their non-CPD counterparts, particularly after the goal-switch event. This proximity to the theoretical limit highlights the CPD module’s ability to promote near-optimal decision sequences by restructuring the temporal hierarchy around true environmental change points.

From a structural standpoint, the improvement can be attributed to the alignment of option boundaries with semantically meaningful transitions. The baseline OC agents, though capable of learning extended temporal abstractions, often misalign option termination points due to reliance on the value-gradient-based termination function $\beta(\omega, s)$. This results in redundant exploration, frequent premature terminations, and suboptimal subgoal transitions. In contrast, the transformer-based CPD component models latent temporal dependencies within the state trajectory and detects potential structural discontinuities in real-time. These detected change points act as supervisory cues for refining $\beta$, enabling terminations to occur precisely when the environment exhibits behavioral or topological discontinuity—such as entering a new room or encountering a new goal configuration. Consequently, the intra-option policies specialize in localized, spatially coherent behaviors, while the high-level policy leverages these temporally aligned segments to make context-aware option selections.



Overall, the empirical evidence suggests that incorporating CPD into the Option-Critic framework results in a significant and sustained improvement in both performance and interpretability. In addition to achieving reduced step counts and quicker recovery following dynamic adjustments, the CPD-augmented agents show more ordered and explicable behavior trajectories. For the deployment of hierarchical reinforcement learning systems in dynamic, real-world environments, the resultant policies must be temporally coherent, hierarchically stable, and closely aligned with the optimum path manifold. These findings support the idea that Transformer-based CPD acts as a potent inductive bias for temporal abstraction, allowing hierarchical agents to learn strong subgoal representations, adaptively break down tasks, and continue to perform close to optimally even in non-stationary scenarios.

\subsection{Pinball Domain Experiment}

In the continuous-control Pinball environment, we evaluated four agent variants: the baseline Option-Critic (OC) with four options, the Option-Critic with Deliberation Cost (OCD) penalizing frequent option switching ($-0.1$ per switch), the proposed CPD-augmented Option-Critic (CPD-OC), and its deliberation-cost counterpart, CPD-OCD, to assess both structural and behavioral effects of integrating CPD and deliberation regularization. 

\begin{figure}
    \centering
    \includegraphics[width=1\linewidth]{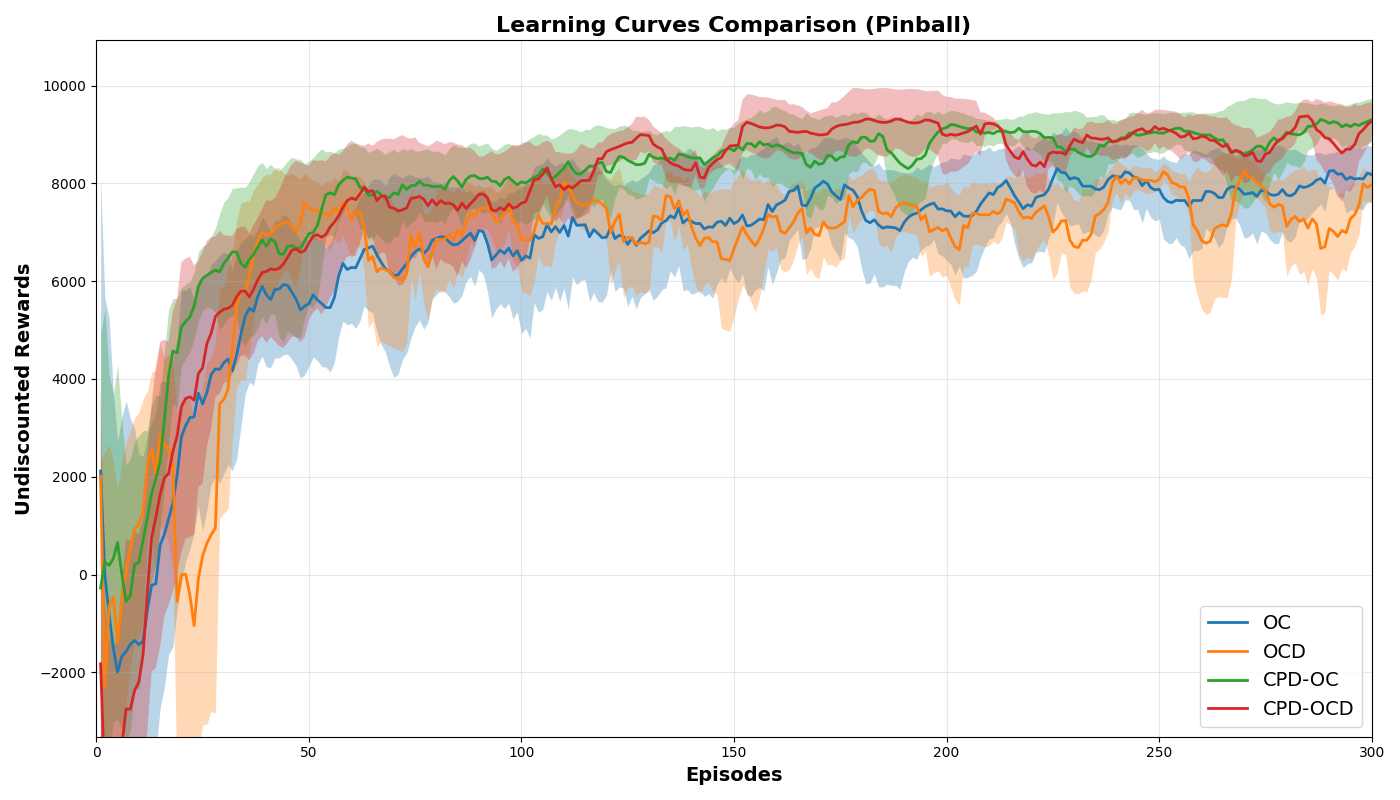}
    \caption{\footnotesize Learning Curve Comparison of Option Critic agent and CPD-enabled Option Critic agents.  
    }
    \label{fig:Pinball Learning Curve}
\end{figure}

As shown in Fig. \ref{fig:Pinball Learning Curve}, both baselines (OC and OCD) converge steadily toward an average reward of $\sim$8500. OCD exhibits smoother learning due to its penalty term, but still plateaus below the optimal region, reflecting the Option-Critic’s lack of explicit state segmentation into coherent subgoals.
However, CPD-OC converges faster, achieves higher asymptotic rewards, and exhibits improved stability. The self-supervised CPD module detects structural discontinuities (e.g., velocity shifts or wall contacts), guiding the termination function $\beta(\omega, s)$ to end options at meaningful moments. This leads to more specialized intra-option policies and better sequencing at the high level.

Quantitatively, CPD-OC reaches a peak reward of $\sim$9500—an 11.8\% improvement over OC—while requiring 30\% fewer episodes. CPD-OCD achieves similar asymptotic returns but with slightly higher variance due to interactions between the deliberation penalty and CPD signals. Both CPD-based agents consistently outperform non-CPD counterparts, confirming that structure-aware temporal segmentation enhances interpretability and efficiency.

In conclusion, both learning speed and ultimate performance are significantly improved by the CPD-OC compared to the original OC and OCD agents. Change point-guided option segmentation transforms the hierarchical policy into a context-sensitive, structure-aware system that can break down complex trajectories into useful behavioral components. The observed performance improvements—faster convergence, higher reward ceilings, and greater policy stability—collectively validate that Transformer-based CPD integration provides a robust inductive bias for temporal abstraction, enabling the agent to align hierarchical decisions with the environment's latent structure. This demonstrates a key advancement in hierarchical reinforcement learning, where option definitions and discoveries are no longer purely static, state-driven, but are informed by learned temporally structured signals from the environment’s underlying dynamics.

\section{DISCUSSION}

These results validate the effectiveness of CPD in time series and how it could enhance learning in HRL.

\subsection{Theoretical Implications}

In this study, a Transformer-based CPD module is integrated into a structure-aware Option-Critic framework for dynamic option termination and discovery. The method reduces vanishing gradients and option collapse by using cross-entropy supervision to align CPD signals with option termination probabilities, resulting in more stable and understandable hierarchical behaviors. CPD-guided behavioral cloning further provides semantically grounded intra-option policies, enabling faster convergence and stronger specialization of subpolicies. Empirically, the proposed CPD-Option-Critic agent achieves substantial performance gains, improving average returns by 32




The proposed CPD-Option-Critic architecture offers a data-driven approach for hierarchical reinforcement learning that generalizes to domains requiring modular and adaptive decision-making. Without environment-specific tuning or manually created subgoals, the agent automatically divides tasks, finds subgoals, and strikes a balance between exploration and exploitation by using Transformer-based change point detection. This makes it ideal for applications such as financial regime analysis, adaptive game AI, and robotic skill segmentation. Beyond performance gains, the framework bridges deep representation learning with interpretable hierarchical control, advancing the development of more robust and generalizable reinforcement learning systems.


\subsection{Limitations and Future Work}

The existing structure has several drawbacks despite its advantages. First, despite the CPD module's strength, its effectiveness is dependent on the caliber of pseudo-label heuristics (such as reward change and TD-error), which can be fragile in highly stochastic contexts or domain-specific. Second, Transformer-based CPD models require sufficient trajectory history to make accurate detections, making the approach potentially less effective in extremely short episodes or online low-data regimes. This work provides a compelling argument for integrating temporal segmentation into hierarchical RL, especially in domains where transition boundaries are latent. By bridging unsupervised change detection and RL, we move towards agents capable of organizing behavior into interpretable and context-aware modules.

In the future, we plan to explore adaptive heuristic learning for pseudo-label generation using meta-learning or unsupervised contrastive objectives. Additionally, we aim to extend this architecture to continuous action spaces and more complex domains, such as robotic manipulation and autonomous driving. Another interesting direction is to incorporate causal inference into the CPD layer to distinguish between agent-induced transitions and environment-induced changes.

\section{Conclusion}

This work presents a novel integration of Transformer-based Change Point Detection (CPD) into the Option-Critic framework, enabling structure-aware option discovery, adaptive termination, and behaviorally consistent policy learning in Hierarchical Reinforcement Learning (HRL). By framing temporal segmentation as a self-supervised sequence modeling problem, the proposed CPD-Option-Critic architecture aligns option boundaries with latent state transitions, effectively bridging time-series segmentation and reinforcement learning.

Our framework leverages CPD outputs to supervise termination functions, initialize intra-option policies, and enforce inter-option diversity through KL-regularized specialization. Empirical results across discrete (Four Rooms) and continuous (Pinball) environments show faster convergence, improved adaptability, and higher cumulative rewards, validating that structural segmentation enhances temporal abstraction and learning efficiency.

Future research will focus on creating adaptable, unsupervised CPD targets, expanding to continuous control tasks, and investigating applications in robotics and time-series forecasting, despite the method having a low computational overhead. All things considered, this study develops a single paradigm for learning temporally organized and interpretable policies for dynamic, complex settings by fusing change point detection with hierarchical reinforcement learning.






\clearpage
\bibliographystyle{ACM-Reference-Format} 
\bibliography{sample}


\end{document}